\pdfoutput=1

\documentclass[11pt]{article}

\usepackage{naacl2021}

\usepackage{times}
\usepackage{latexsym}

\usepackage[T1]{fontenc}

\usepackage[utf8]{inputenc}

\usepackage{microtype}
\usepackage{hhline}

\usepackage{adjustbox}

\usepackage[ruled,vlined]{algorithm2e}

\usepackage{setspace}
\usepackage{scrextend}
\usepackage{mathtools}

\DeclarePairedDelimiter\floor{\lfloor}{\rfloor}

\makeatletter
\newcommand\footnoteref[1]{\protected@xdef\@thefnmark{\ref{#1}}\@footnotemark}
\makeatother

\usepackage{microtype}

\usepackage{smile}

\usepackage{todonotes}

\title{Token-wise Curriculum Learning for Neural Machine Translation}

\author{\makecell{Chen Liang\thanks{~~Work was done at Microsoft Dynamics 365 AI.}~~$^\diamond$, Haoming Jiang$^\diamond$, Xiaodong Liu$^\dagger$, Pengcheng He$^\ddagger$, \\
Weizhu Chen$^\ddagger$, Jianfeng Gao$^\dagger$ and Tuo Zhao$^\diamond$} \\
$^\diamond$ Georgia Tech, $^\dagger$ Microsoft Research, $^\ddagger$ Microsoft Dynamics 365 AI
} 

\begin{document}
\maketitle


\begin{abstract}


Existing curriculum learning approaches to Neural Machine Translation (NMT) require sampling sufficient amounts of ``easy'' samples from training data at the early training stage. This is not always achievable for low-resource languages where the amount of training data is limited. To address such limitation, we propose a novel token-wise curriculum learning approach that creates sufficient amounts of easy samples. Specifically, the model learns to predict a short sub-sequence from the beginning part of each target sentence at the early stage of training, and then the sub-sequence is gradually expanded as the training progresses. Such a new curriculum design is inspired by the cumulative effect of translation errors, which makes the latter tokens more difficult to predict than the beginning ones. Extensive experiments show that our approach can consistently outperform baselines on 5 language pairs, especially for low-resource languages. Combining our approach with sentence-level methods further improves the performance on high-resource languages. 


\end{abstract}

\section{Introduction}
\label{sec:intro}

Neural Machine Translation (NMT) has achieved significant progress in recent years \citep{sutskever2014sequence, bahdanau2014neural, vaswani2017attention}, mainly in the scenarios where the parallel training corpora are abundant. However, training corpora can be limited in some domains (e.g., spoken language \citep{cettolo2015iwslt}) and languages (e.g., African languages) due to the high cost of data acquisition. \citet{koehn2017six, lample2018phrase} show that NMT models do not perform well in such data-limited settings.

To improve NMT with limited data, researchers  resort to large amounts of auxiliary data. One line of research leverages the knowledge from high-resource parallel corpora. For examples, some works pre-train  NMT models on high-resource data, and then fine-tune them on low-resource data \citep{zoph2016transfer, chen2017teacher, kocmi2018trivial, neubig2018rapid, nguyen2017transfer}; others train Multilingual or Multitask NMT models jointly on both high-resource and low-resource datasets \citep{gu2018universal, gu2018meta, aharoni2019massively, jiang2019multi, siddhant2020leveraging}. 
The other line exploits high-resource monolingual data as auxiliary data to train NMT models in a semi-supervised manner \citep{sennrich2015improving, currey2017copied, cheng2019semi}. 

Aside from previous approaches, curriculum learning \citep{bengio2009curriculum} is proposed to address the data insufficiency issue via utilizing the limited data more efficiently \citep{zhang2019curriculum}. The idea of curriculum learning is to sample training data in an order of increasing difficulty. The ``easy'' samples of such a curriculum can be beneficial to the training of the models at the early stage. There have been multiple designs of curriculum for NMT in the recent literatures \cite{zhou2020uncertainty, liu2020norm, ruiter2020self, platanios2019competence, wang2019dynamically, wang2019learning, kumar2019reinforcement, zhang2018empirical}. 
All these methods sample complete sentence pairs for training from a selected \textit{subset}, which expands as training progresses. We refer to them as the ``sentence-level'' curricula. 

\begin{figure}[!ht]
    \centering
    \adjustbox{trim={0.0\width} {0.04\height} {0.\width} {0.02\height},clip}{
    \includegraphics[width=1\linewidth]{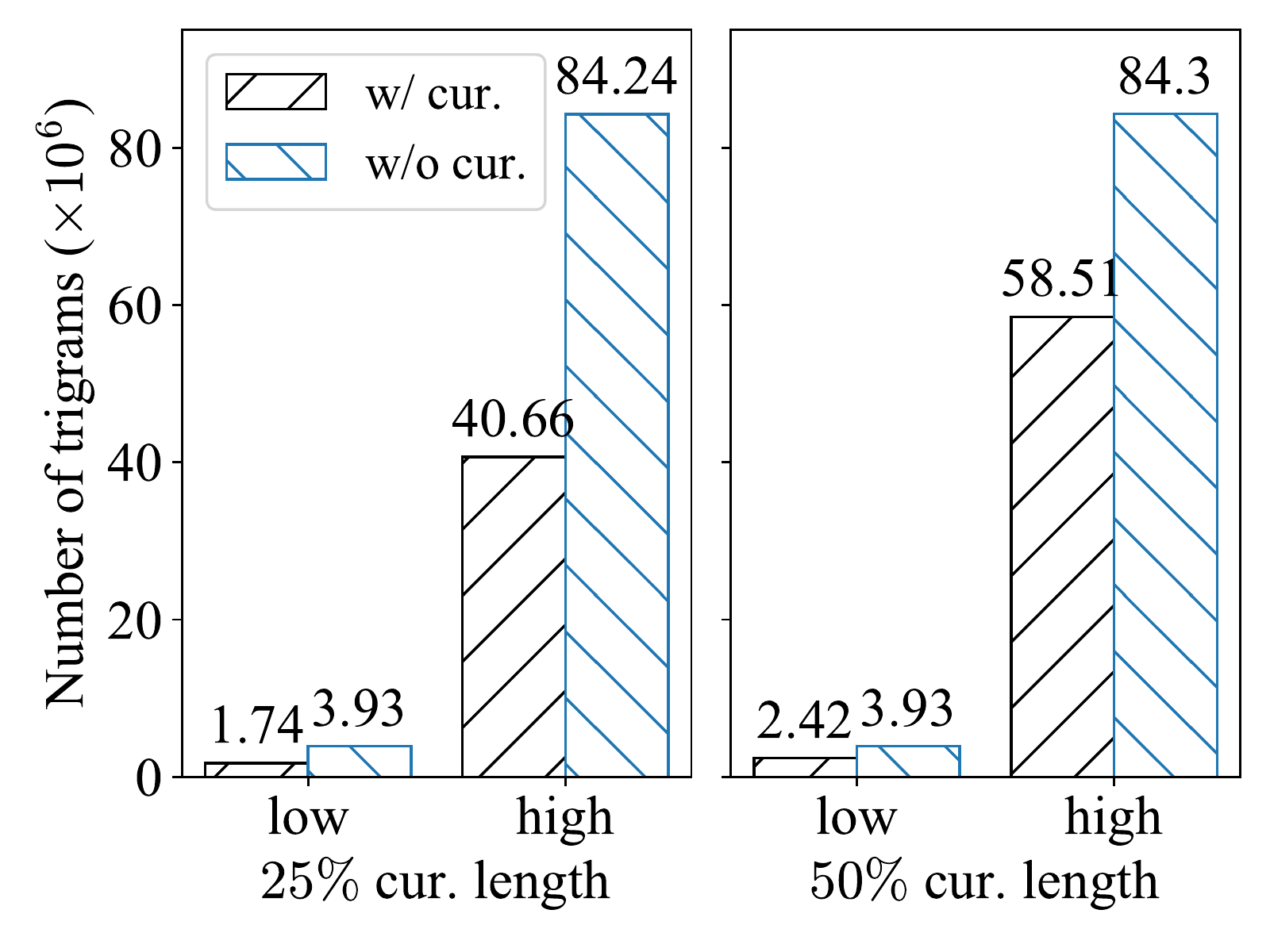}
    }
	\caption{Sample diversity in early stage of training under low-resource and high-resource settings.}
	\label{fig:diversity}
\end{figure}

\begin{figure}[!ht]
    \centering
    \vspace{-0.2in}
    \adjustbox{trim={0.0\width} {0.02\height} {0.\width} {0.01\height},clip}{
    \includegraphics[width=1\linewidth]{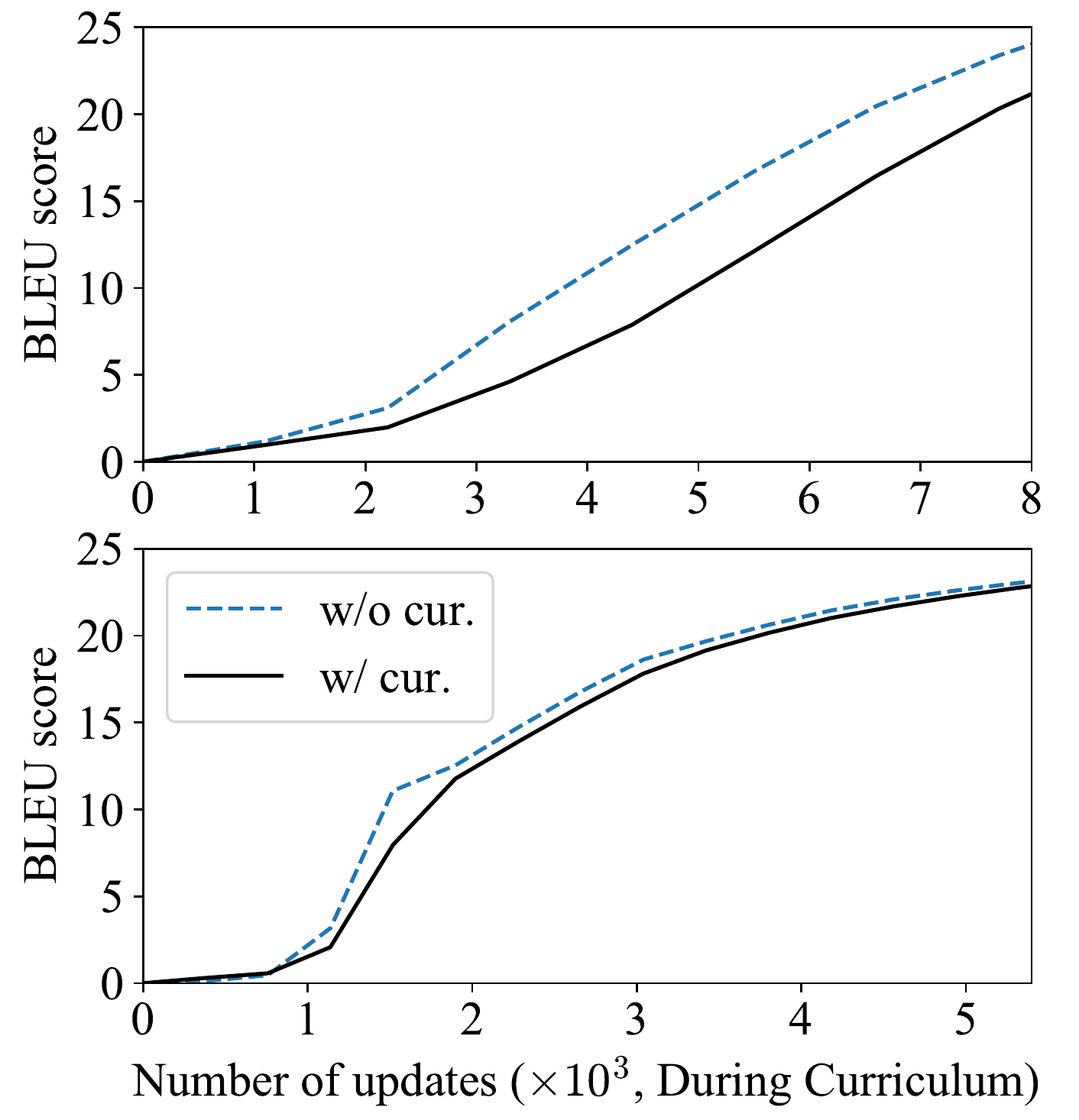}
    }
	\caption{BLEU score (dev) in low-resource (upper) and high-resource (lower) settings.}
	\label{fig:low_vs_high}
\end{figure}

However, such a sentence-level design is not necessarily effective for NMT when data is limited. In the early stage of training, the selected \textit{subset} is usually limited to a small portion of total training samples. In the low-resource setting, this \textit{subset} contains even fewer samples. To better measure this effect, we use Figure~\ref{fig:diversity} to show the diversity of the samples selected in the early training stage under low-resource and high-resource settings \footnote{For all sentence-level curriculum experiments in Section~\ref{sec:intro}, we adopt the design proposed in \citet{zhou2020uncertainty}. The dataset we use in low-resource setting is IWSLT14 De-En, and in high-resource setting is WMT16 En-De. We adopt Transformer-base \citep{vaswani2017attention} as the baseline model.}. Specifically, we count the number of unique trigrams in the sentence pairs 
used for training up-till a certain training iteration.
We observe that the selected samples in low resource setting are less diverse than in high-resource setting, especially in the early curriculum (i.e. up-till 25\% of total updates in curriculum). Consequently, the sentence-level curriculum slows down the learning progress in low-resource setting although this is not an issue for high-resource setting, as shown in Figure~\ref{fig:low_vs_high}. 

This observation, that insufficient diversity in low-resource setting can affect the learning efficiency, motivates us to design a token-wise curriculum. During the curriculum, the model learns to predict only a short sub-sequence from $\textit{each}$ target sentence at the early stage of training, and then the sub-sequence is gradually expanded as the training progresses. 
Comparing with the sentence-level curriculum, which only focuses on ``easy'' sentence pairs, token-wise curriculum can on purpose create much more partial and diverse samples to address the data insufficiency challenge.

The next question is, how to design an effective sub-sequence selection scheme, such that the difficulty of the selected sub-sequences follows an ``easy-to-hard'' schedule. Specifically, we consider the sub-sequence difficulty in the context of machine translation generation. In a left-to-right autoregressive generation setting, the generation of next word is dependent on previous generation on the left. In other words, wrong predictions in the early tokens would affect the accuracy of the latter ones during inference. This results in prediction error accumulation\footnote{We verify that the phenomenon of error accumulation exists in NMT in Section~\ref{sec:nmt_ablation} Figure~\ref{fig:pre_ea}.} \citep{zhang2019bridging}, which indicates that the beginning tokens are easier to predict than the latter ones. Therefore, we design a scheduler to select sub-sequences from the $\textit{beginning}$ part of target sentences, and gradually expand them until the end of the sentences, as the training progresses. 




Our experiments on several low-resource NMT datasets collected from IWSLT \citep{cettolo2015iwslt} show that the proposed curriculum outperforms existing baselines in both standard training and transfer learning settings. In addition, the experiments on a high-resource dataset WMT'16 En-De \citep{bojar2016findings} shows that the proposed curriculum can not only by itself, but also by combining with existing sentence-level curricula, benefit NMT model training. Finally, we show that the proposed token-wise curriculum is general for multiple sequence generation tasks and we show its superior performance on language modeling tasks, besides machine translation. Our codes are released at \url{https://github.com/cliang1453/token-level-curriculum-learning}.

\section{Background}
\label{sec:related}

\noindent $\bullet$ \textbf{NMT} models the  conditional probability of a target sentence $\boldsymbol{y} = (y_1, ..., y_{\ell})$ given a source sentence $\boldsymbol{x} = (x_1, ..., x_{m})$. The density function $p(\boldsymbol{y}|\boldsymbol{x})$ is parameterized by an encoder-decoder neural network, which generates the target sentence in an auto-regressive manner \citep{sutskever2014sequence, bahdanau2014neural}. Specifically, the model predicts the probability of the $t$-th token by $p(y_{t}|y_{<t}, \boldsymbol{x}; \theta)$, where $\theta$ denotes the model parameters. It is trained by minimizing the sum of cross-entropy loss on all sentence pairs, where the loss on each sentence pair $(\boldsymbol{x}, \boldsymbol{y})$ is
\begin{align}
    L(\boldsymbol{x}, \boldsymbol{y}; \theta) = -\frac{1}{\ell}\sum_{t=1}^{\ell} \log p(y_{t}|y_{<t}, \boldsymbol{x}; \theta)
\end{align}

\noindent $\bullet$ \textbf{Curriculum Learning in NMT}. Research on curriculum learning in NMT mainly fall into two categories: measurement of sample difficulty and design of curriculum schedule \citep{kocmi2017curriculum}. In the first category, some research measure sample difficulty with features derived from lexical statistics, e.g., sentence length and word rarity \citep{zhang2018empirical, platanios2019competence}. Others measure difficulty with features derived from pre-trained model, e.g., \citet{liu2020norm} use the norm of pre-trained word embeddings, \citet{wang2018denoising} leverage a pre-trained NMT model to measure sample noise-level, and \citet{zhou2020uncertainty} use a pre-trained language model to measure the word-level perplexity (i.e. uncertainty). In the second category, most schedules select samples with difficulty under a threshold. \citet{platanios2019competence} determine the threshold by a linear/square-root function of training step, \citet{liu2020norm} design a function based on the norm of the encoder word embedding, and \citet{zhou2020uncertainty} design a function based on model uncertainty.

\section{Method}
\label{sec:method}
We introduce a token-wise curriculum learning approach for NMT.

\subsection{Hard Curriculum}
\label{sec:hard}
We propose a token-wise curriculum based on sub-sequence selection. At each training step, the model is trained to predict a \textit{sub-sequence} of target sentences. We remark that such prediction is conditioned on \textit{complete} source sentences. Specifically, the model is updated based on the loss computed on such sub-sequence only, 
\begin{align}
     -\frac{1}{|S_i|}\sum_{t \in S_i} \log p(y_{t}|y_{<t}, \boldsymbol{x}; \theta),
\end{align}
where $S_i$ is the set of the token indexes in the selected sub-sequence at the $i$-th iteration.


\noindent\textbf{Left-to-Right Selection Scheme.} The selection scheme of $S_i$ can be described as follows: 

\noindent$\bullet$ At the beginning of curriculum ($0$-th iteration), we select the sub-sequence from the beginning of each target sentence:  $S_0 = [1,2,3,...,\floor{\lambda_0\ell}]$, where $\ell$ is the length of the target sequence and $\lambda_0 \in (0,1)$ is the initial sub-sequence percentage with respect to the total length.

\noindent$\bullet$ We then gradually expand each sub-sequence throughout the curriculum until it covers the whole sentence: $S_i = [1,2,3,...,\ell_i]$ with the length $\ell_i$ determined by a linear function, 
\begin{align}
    \ell_i &= \floor{\ell \cdot (\lambda_0 + \frac{i}{I} \cdot (1-\lambda_0))} \quad 0 < i < I,
\end{align}
where $I$ is the number of updates in the curriculum. 

With this selection scheme, the model can be updated by the SGD type algorithm (e.g., ADAM \citep{kingma2014adam}) with the stochastic gradients computed based on $S_i$. The gradient of each sentence is computed by:
\begin{align}
    \frac{1}{\ell_{i}} \sum_{t=1}^{\ell_{i}} \nabla_{\theta} -\log p(y_{t}|y_{<t}, \boldsymbol{x}; \theta). 
    \label{equ:hard}
\end{align}
After the curriculum ends (i.e., $i \geq I$), the model continues with the standard training. 

\subsection{Soft Curriculum}
\label{sec:soft}

In hard curriculum, the model is trained without regarding the loss upon $\{1..\ell\} \setminus S_i$. However, those tokens may play important roles in sequence generation. For example, the model needs to learn how to end a sentence by predicting the $\langle EOS \rangle$ token. Therefore, we propose an alternative method -- the soft curriculum, where we place geometrically decaying weights on the loss of all tokens. By allowing weights on all tokens, the model is able to learn more diverse samples. By placing decaying weight on end tokens that are difficult to learn, we maintain the sample easiness. 

\begin{figure}[!ht]
    \centering
    \adjustbox{trim={0.0\width} {0.05\height} {0.\width} {0.02\height},clip}{
    \includegraphics[width=1.0\linewidth]{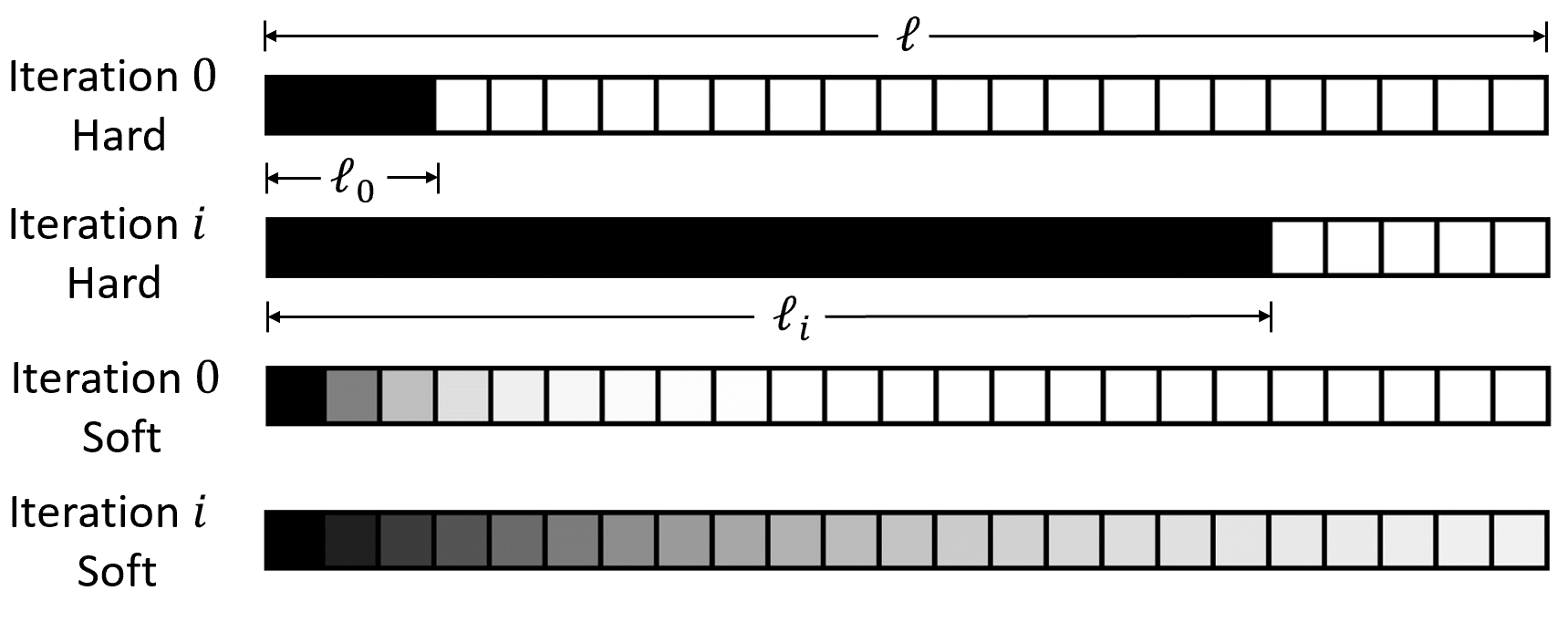}
    }
	\caption{A comparison between sub-sequence selection in hard and soft curricula. Each row of grids represents a target sentence of tokens. The depth of color from light$\xrightarrow{}$dark represents the weight on the token loss from 0$\xrightarrow{}$1. }
	\label{fig:hard_vs_soft}
	\vspace{-5mm}
\end{figure}

At the $i$-th iteration, the re-weighted stochastic gradients on each target sentence with length $\ell$ is computed by:
\begin{align}   
    \frac{1}{\ell}\sum_{t=1}^{\ell} \gamma_i ^ {\alpha_{t, \ell}} \cdot \nabla_{\theta} -\log p(y_{t}|y_{<t}, \boldsymbol{x}, \theta),
    \label{equ:soft}
\end{align}
where $\gamma_i$ and $\alpha_{t, \ell}$ are two factors controlling the rate of geometric decay. 
The decaying factor $\gamma_i$ at the $i$-th iteration is computed by:
\begin{align}
    \gamma_i = \gamma_0 + \frac{i}{I} \cdot (1 - \gamma_0) \quad 0 < i < I,
    \label{equ:soft_decay}
\end{align}
where $0 \leq \gamma_0 < 1$ is a hyperparameter controlling the scale of initial weights placed on all tokens. The weights gradually increase as $\gamma_i$ grows from $\gamma_0$ to $1$ throughout the curriculum. We remark that while $\gamma_i$ grows linearly, the weights change with different rates for tokens at different positions -- we design the power factor $\alpha_{t, \ell}$ uniquely for the $t$-th token in a target sentence of length $\ell$:
\begin{align}
    \alpha_{t, \ell} = \alpha_{0} \cdot \frac{t-1}{\ell-1}& \quad 0 < t \leq \ell.
    \label{equ:soft_power}
\end{align}
As illustrated in Figure~\ref{fig:hard_vs_soft}, the weights on tokens gradually decay from the beginning to the end of the sentence, where $\alpha_0 > 0$ is a hyperparameter controlling this decaying rate. 

\section{NMT Experiments}
\label{sec:nmt_exp}
To demonstrate the effectiveness of our token-wise curriculum design, we present experimental results on NMT tasks.

\subsection{Data Preparation \& Preprocessing}
\label{sec:nmt_data}
We evaluate our method on widely used language pairs in both low-resource and high-resource datasets. 
Low-resource datasets include English-to-Vietnamese (En-Vi) from IWSLT15 \citep{cettolo2015iwslt}\footnote{\small{\url{https://wit3.fbk.eu/}}\label{iwslt}}, German-to-English (De-En) from IWSLT14, French-to-English (Fr-En) from IWSLT16, and Romanian-to-English (Ro-En) from WMT16 \citep{bojar2016findings}\footnote{\small{\url{http://data.statmt.org/wmt16/translation-task/}}\label{wmt}}. 
The high-resource dataset is the WMT16 En-De. Table~\ref{tab:data} shows the number of sentence pairs in each dataset. See~\ref{app:datasets} for details on dev and test set used.


\begin{table}[!ht]
\centering
\begin{tabular}{@{~}l@{~~~}c@{~~~}c@{~~~}c@{~~~}c@{~}}
\hline 
\textbf{Data}  &  \textbf{Train} & \textbf{Dev} & \textbf{Test} \\ \toprule 
\textbf{En-Vi} &  133K & 768   & 1268 \\
\textbf{De-En} &  160K & 7283  & 6750 \\
\textbf{Fr-En} &  224K & 1080  & 1133 \\
\textbf{Ro-En} &  612K & 1999  & 1999 \\ \hline
\textbf{En-De} &  4.5M & 1061  & 1019 \\
\bottomrule
\end{tabular}
\caption{\label{tab:data} The number of parallel sentences in datasets.}
\end{table} 

All datasets are encoded using byte-pair encoding (BPE, \citet{sennrich2016edinburgh}). For En-Vi and Fr-En, we use a BPE trained with $32$K merge operations and use sentences up to length $200$ subword symbols, following \citet{platanios2019competence}. For Ro-En, we use a BPE trained with $40$K merge operations and use sentences up to length $50$ subword symbols as \citet{gu2018universal, gu2018meta}. We preprocess De-En data following \textit{fairseq}\footnote{\small{\url{https://github.com/pytorch/fairseq/blob/master/examples/translation/prepare-iwslt14.sh}}\label{fairseq}}. We adopt the preprocessed En-De data released by Google\footnote{\small{\url{https://pytorchnlp.readthedocs.io/en/latest/_modules/torchnlp/datasets/wmt.html}}}.
\subsection{Baselines}
\label{sec:nmt_baselines}
We compare our token-wise curriculum learning method ($\textbf{TC}$) with several state-of-the-art sentence-level methods ($\textbf{SC}$): 

\noindent $\bullet$ $\textbf{SC}_{\text{r-sqrt}}$ measures sample difficulty by word rarity, and uses a square-root function as curriculum schedule \citep{platanios2019competence}. \\
\noindent $\bullet$ $\textbf{SC}_{\text{norm}}$ measures sample difficulty based on norm of sentence embedding, and uses a threshold function of encoder word embedding norm as curriculum schedule \citep{liu2020norm}. \\
\noindent $\bullet$ $\textbf{SC}_{\text{unc}}$ measures sample difficulty by data uncertainty, and uses a threshold function of model uncertainty as curriculum schedule \citep{zhou2020uncertainty}.

\subsection{Model \& Training}
\label{sec:nmt_model}

For both $\textbf{SC}$ and $\textbf{TC}$ experiments, we adopt Transformer-base NMT model \citep{vaswani2017attention} as the baseline model. All implementations are based on \textit{fairseq} \citep{ott2019fairseq} code-base with all experiments running with $32$G NVIDIA V100 GPUs. For all datasets, we use ADAM \citep{kingma2014adam} as the optimizer with $\beta=(0.9, 0.98)$. For low-resource datasets, we use a learning rate of $5\times10^{-4}$ with $8000$ steps of warmup updates. For high-resource dataset En-De, we use a learning rate of $1\times10^{-3}$ with $4000$ steps of warmup updates. See training details in~\ref{app:training}.

We fix $\lambda_0 = 0.1$ in $\textbf{TC}_{\text{hard}}$ experiments, and fix $\gamma_0 = 0.7$ and $\alpha_0 = 25$ in $\textbf{TC}_{\text{soft}}$ experiments. We set the curriculum length $I = 8000, 7000, 6500, 1100, 5400$ for De-En, En-Vi, Fr-En, Ro-En and En-De. See hyperparameter selection details in~\ref{app:param_study}. For $\textbf{SC}$ methods, we follow the recommended settings in the original papers with special configurations for the low-resource setting. See training details in~\ref{app:baseline}.

We use BLEU \citep{papineni2002bleu} as the evaluation metric. For all low-resource datasets, we report the BLEU score of the best checkpoint using a beam size of $5$ and length penalty of $1$. For high-resource dataset En-De, we report the average of the last $10$ checkpoints with a beam size of $10$ and length penalty of $0.6$.

\subsection{Main Results}
\label{sec:nmt_main}
We compare $\textbf{TC}_{\text{hard}}$ and $\textbf{TC}_{\text{soft}}$ with the baseline, and report the best testing BLEU among 5 runs with different random seeds in Table~\ref{tab:main_nmt_low} and Table~\ref{tab:main_nmt_high} (See~\ref{app:valid} for validation scores). As can be seen, $\textbf{TC}_{\text{hard}}$ outperforms the baseline in all cases, and $\textbf{TC}_{\text{soft}}$ further improves upon $\textbf{TC}_{\text{hard}}$. This implies that $\textbf{TC}_{\text{soft}}$ finds a better balance between sample diversity and sample easiness than $\textbf{TC}_{\text{hard}}$. 

In the low resource setting (Table~\ref{tab:main_nmt_low}), all $\textbf{TC}$ methods uniformly outperform $\textbf{SC}$ methods by around $0.5$ BLEU scores, while $\textbf{SC}$ methods can sometimes hurt the baseline (e.g., in En-Vi and De-En, the two smallest datasets). Under the high resource setting (Table~\ref{tab:main_nmt_high}), all $\textbf{TC}$ methods outperform the baseline by around $0.4$ BLEU scores. However, we observe that the performance of $\textbf{TC}$ methods show no clear improvement upon $\textbf{SC}$ methods. We conjecture the reason is that the selected samples in high resource setting are sufficiently diverse for $\textbf{SC}$ method to be well-performed. To further improve performance in high resource setting, we combine $\textbf{TC}$ and $\textbf{SC}$ methods, expecting that this combination selects not only diverse, but also easier samples than those selected by any single method. In particular, we first use $\textbf{SC}$ to select sentences, and then use $\textbf{TC}$ to select beginning sub-sequences upon these sentences. As can be seen, both $\textbf{TC}_{\text{soft}}+\textbf{SC}_{\text{norm}}$ and $\textbf{TC}_{\text{soft}}+\textbf{SC}_{\text{unc}}$ can further improve upon the best single method.


As $\textbf{TC}_{\text{soft}}$ uniformly outperforms $\textbf{TC}_{\text{hard}}$, we use $\textbf{TC}_{\text{soft}}$ in the following experiments unless stated otherwise.  

\begin{table}[!htb]
        \centering
        \begin{tabular}{l|cccc}
        \toprule
                                      & \textbf{En-Vi}       & \textbf{De-En}       & \textbf{Fr-En}       & \textbf{Ro-En} \\ \midrule
        w/o Cur.                       & 31.43       & 34.33       & 37.21       & 32.10       \\ 
        $\textbf{SC}_{\text{r-sqrt}}$          & 31.01	    & 34.29       & 37.25	    & 32.19       \\
        $\textbf{SC}_{\text{norm}}$            & 31.05      & 34.24       & 37.28       & 32.25       \\
        $\textbf{SC}_{\text{unc}}$          & 31.33	    & 34.48       & 37.60	    & 32.26       \\ \midrule
        $\textbf{TC}_{\text{hard}}$            & 31.85      & 34.88       & 38.22       & 32.45       \\
        $\textbf{TC}_{\text{soft}}$            & \textbf{31.94}       & \textbf{34.91}       & \textbf{38.28}       & \textbf{32.52}\\ \bottomrule
        \end{tabular}
        \caption{\label{tab:main_nmt_low} BLEU scores (test) on low-resource datasets. The performance gains of $\textbf{TC}_{\text{hard}}$ and $\textbf{TC}_{\text{soft}}$ are significant compared with $\textbf{SC}_{\text{unc}}$ -- the results of $5$ runs pass the unpaired student's $t$-test with p-value $< 0.01$.}
\end{table}

\begin{table}[!htb]
        \centering
        \begin{tabular}{l|c}
        \toprule
                                             & \textbf{En-De} \\ \midrule 
        w/o Curriculum  \citep{vaswani2017attention}                & 28.10                    \\ 
        $\textbf{SC}_{\text{r-sqrt}}$   \citep{platanios2019competence}     & 28.27                    \\
        $\textbf{SC}_{\text{norm}}$     \citep{liu2020norm}     & 28.51                     \\
        $\textbf{SC}_{\text{unc}}$   \citep{zhou2020uncertainty}     & 28.55                      \\ \midrule
        $\textbf{TC}_{\text{hard}}$          & 28.49                      \\
        $\textbf{TC}_{\text{soft}}$          & 28.54 \\ 
        $\textbf{TC}_{\text{soft}} + \textbf{SC}_{\text{norm}}$     & 28.62 \\ 
        $\textbf{TC}_{\text{soft}} + \textbf{SC}_{\text{unc}}$      & \textbf{28.67} \\ \bottomrule
        \end{tabular}
        \caption{\label{tab:main_nmt_high} BLEU scores (test) on the high-resource datasets. The performance gains of $\textbf{TC}_{\text{soft}}$, $\textbf{TC}_{\text{hard}}$, $\textbf{TC}_{\text{soft}} + \textbf{SC}_{\text{norm}}$ and $\textbf{TC}_{\text{soft}} + \textbf{SC}_{\text{unc}}$ are significant compared with w/o Curriculum -- the results of $5$ runs pass the unpaired student's $t$-test with p-value $< 0.01$.}
\vspace{-4mm}
\end{table}


\subsection{Transfer Learning with Curriculum}
\label{sec:nmt_trans}

We show that our curriculum can be further combined with transfer learning to improve NMT performance. 
Instead of training from scratch, transfer learning considers fine-tuning a pre-trained model on the limited parallel data.
Specifically, we consider the following transfer learning settings:

\noindent $\bullet$ \textbf{Domain Transfer Learning}. We consider transferring from a high-resource domain to a low-resource domain. Specifically, we fine-tune the Transformer-big NMT model pre-trained from News domain (WMT)  \footnote{The pre-trained model is trained on WMT16 En-De data and publicly available from \small{\url{github.com/pytorch/fairseq/tree/master/examples/translation}}. } on TED domain (IWSLT). Table~\ref{tab:transfer} shows that using our curriculum improves the domain transfer performance. 
    
\noindent $\bullet$ \textbf{Pre-trained Multilingual Language Model Fine-tuning}. We also consider the case of transferring from high-resource monolingual data to low-resource parallel data. Specifically, we initialize an NMT model from XLM \citep{lample2019cross}, a multilingual language model pre-trained on extensive monolingual En and De data \footnote{The pre-trained XLM model and script for fine-tuning translation models are publicly available \small{\url{github.com/facebookresearch/XLM}}.}. 
Then we fine-tune the NMT model on the En-De TED data. Table~\ref{tab:transfer} shows that using our curriculum improves the fine-tuning performance on pre-trained multilingual language model. 
    


\begin{table*}[!htb]
\centering
\begin{tabular}{l|cccc|cccc} \toprule
    Pre-trained Model & \multicolumn{4}{c|}{Translation Model}               & \multicolumn{4}{c}{Multilingual Language Model (XLM)}           \\ \hline
    Source            & \multicolumn{4}{c|}{\textbf{En-De News} (Parallel Data)}  & \multicolumn{4}{c}{\textbf{En-De News} (Monoligual Data)}   \\
    Target            & \multicolumn{4}{c|}{\textbf{En-De TED} }  & \multicolumn{4}{c}{\textbf{En-De TED} } \\ \hline
    Size & full & 50\% & 10\% & 1\% & full & 50\% & 10\% & 1\% \\ \midrule
    Transfer w/o Curriculum & 32.88 & 32.42 & 31.30 & 26.66  & 31.86 & 28.22 & 14.52 & 9.15 \\
    \textbf{TC}$_{\text{soft}}$& \textbf{33.17} & \textbf{32.79} & \textbf{31.77} & \textbf{29.69} & \textbf{33.22} & \textbf{29.94} & \textbf{17.17} &  \textbf{10.75} \\ \bottomrule 
\end{tabular}
\caption{BLEU scores (test) in transfer learning, finetuned with full/$50\%$/$10\%$/$1\%$ subsets of target domain data.}
\label{tab:transfer}
\end{table*}


\subsection{Curriculum under Extremely Low-Resource Setting}
\label{sec:nmt_extreme}
We further show that our curriculum can improve NMT performance in both standard training (Table~\ref{tab:extreme_low}) and transfer learning (Table~\ref{tab:transfer}) under extremely low-resource setting.  In standard training, the model is trained with a randomly sampled $50\%$/$10\%$ subset from all sentence pairs. In transfer learning, the model is finetuned with a randomly sampled $50\%$/$10\%$/$1\%$ subset from all target domain sentence pairs. Table~\ref{tab:transfer} and Table~\ref{tab:extreme_low} show that $\textbf{TC}_{\text{soft}}$ attains a steady performance gain as training/fine-tuning data becomes more scarce, e.g., the domain transfer learning improvement is over $3$ BLEU scores under the $1\%$ data setting.

\begin{table*}[!ht]
\centering
\begin{tabular}{l|ccc|ccc|ccc}
\hline
                                  & \multicolumn{3}{c|}{\textbf{De-En}} & \multicolumn{3}{c|}{\textbf{Fr-En}} & \multicolumn{3}{c}{\textbf{Ro-En}} \\ \toprule 
Size                                  & full      & 50\%      & 10\%     & full      & 50\%      & 10\%     & full      & 50\%      & 10\%     \\ \midrule 
w/o Curriculum                     & 34.33     & 31.04	  & 16.33    & 37.21     & 34.25	 & 21.36    & 32.10     & 29.99	    & 21.96    \\
$\textbf{TC}_{\text{soft}}$           & \textbf{34.91}     & \textbf{31.55}     & \textbf{16.83}    & \textbf{38.28} & \textbf{34.66}     & \textbf{21.62}    & \textbf{32.52}     & \textbf{30.22}    & \textbf{22.25}    \\ \bottomrule
\end{tabular}
\caption{\label{tab:extreme_low} BLEU scores (test) on NMT, trained with full/$50\%$/$10\%$ subsets of training sentence pairs.}
\end{table*}

\subsection{Analysis}
\label{sec:analysis}

We first verify our assumption that the error accumulation makes beginning tokens easier to predict. Then we analyze whether our curriculum improves the sample diversity in the early stage of training, and further improves optimization. 

\begin{figure}[!ht]
    \centering
    \vspace{-0.3in}
    \includegraphics[width=1.05\linewidth]{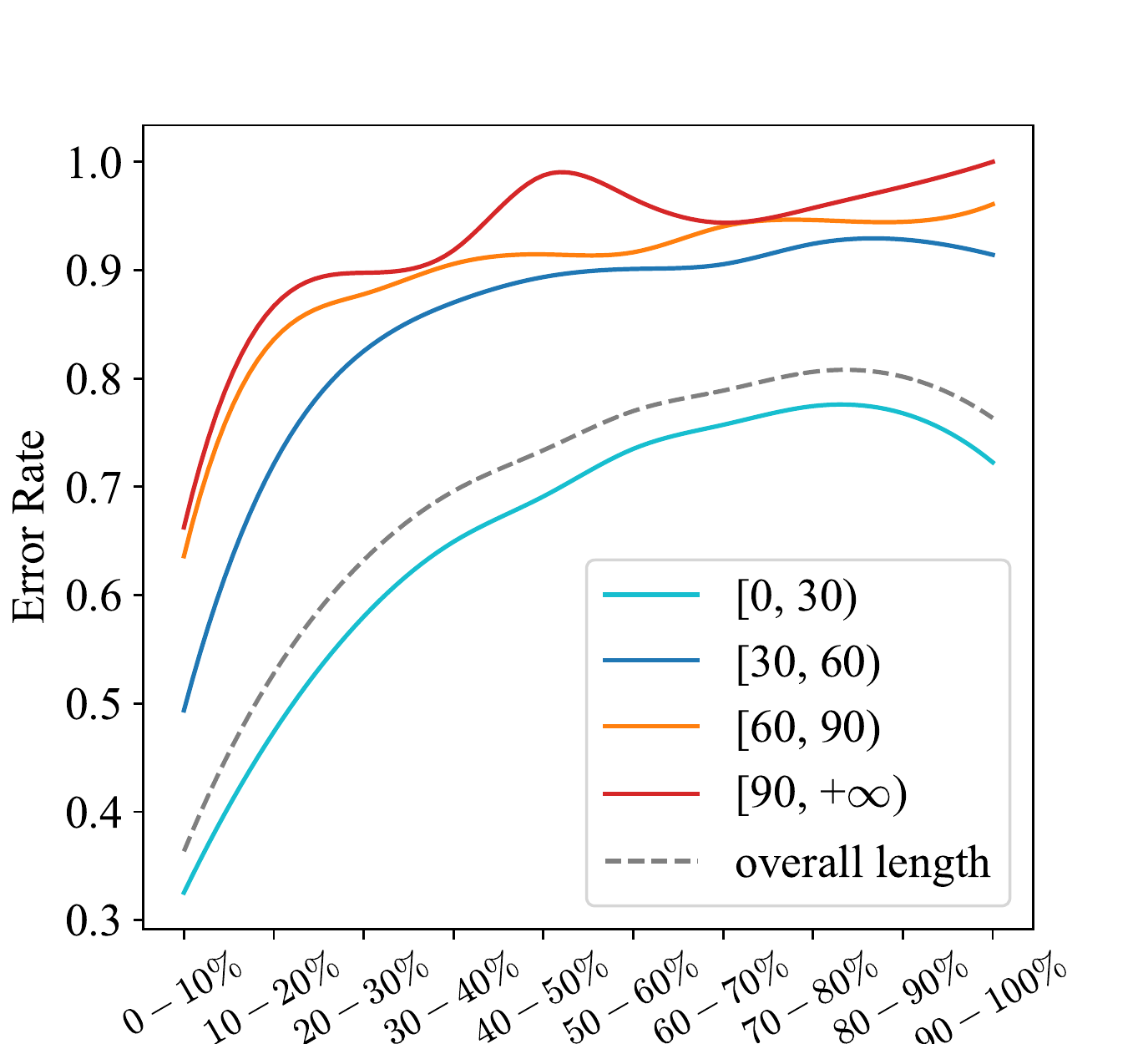}
    \vspace{-0.2in}
	\caption{The averaged beam search error rate at different relative positions of sentences.} 
	\label{fig:pre_ea}
	\vspace{-0.2in}
\end{figure}

\noindent $\bullet$ \textbf{Error Accumulation}. To verify that error accumulation is a prevailing phenomenon in machine translation generation, we conduct beam search with beam size of $5$ using Transformer-base NMT model on De-En dataset. We compute the error rate of the predictions at different relative positions of sentences. Specifically, we compute the prediction error rate within $10$ evenly-divided partitions in each sentence and average over all sentences. Since we choose $\lambda_0$ invariant to sentence length, we further verify that the error accumulation exists for sentences with different length. As shown in Figure~\ref{fig:pre_ea}, sentences with different length suffer from error accumulation. See more details in~\ref{app:add_ana}. 

\begin{figure}[!ht]
    \vspace{-0.01in}
    \centering
    \adjustbox{trim={0.0\width} {0.03\height} {0.\width} {0.1\height},clip}{
    \includegraphics[width=1\linewidth]{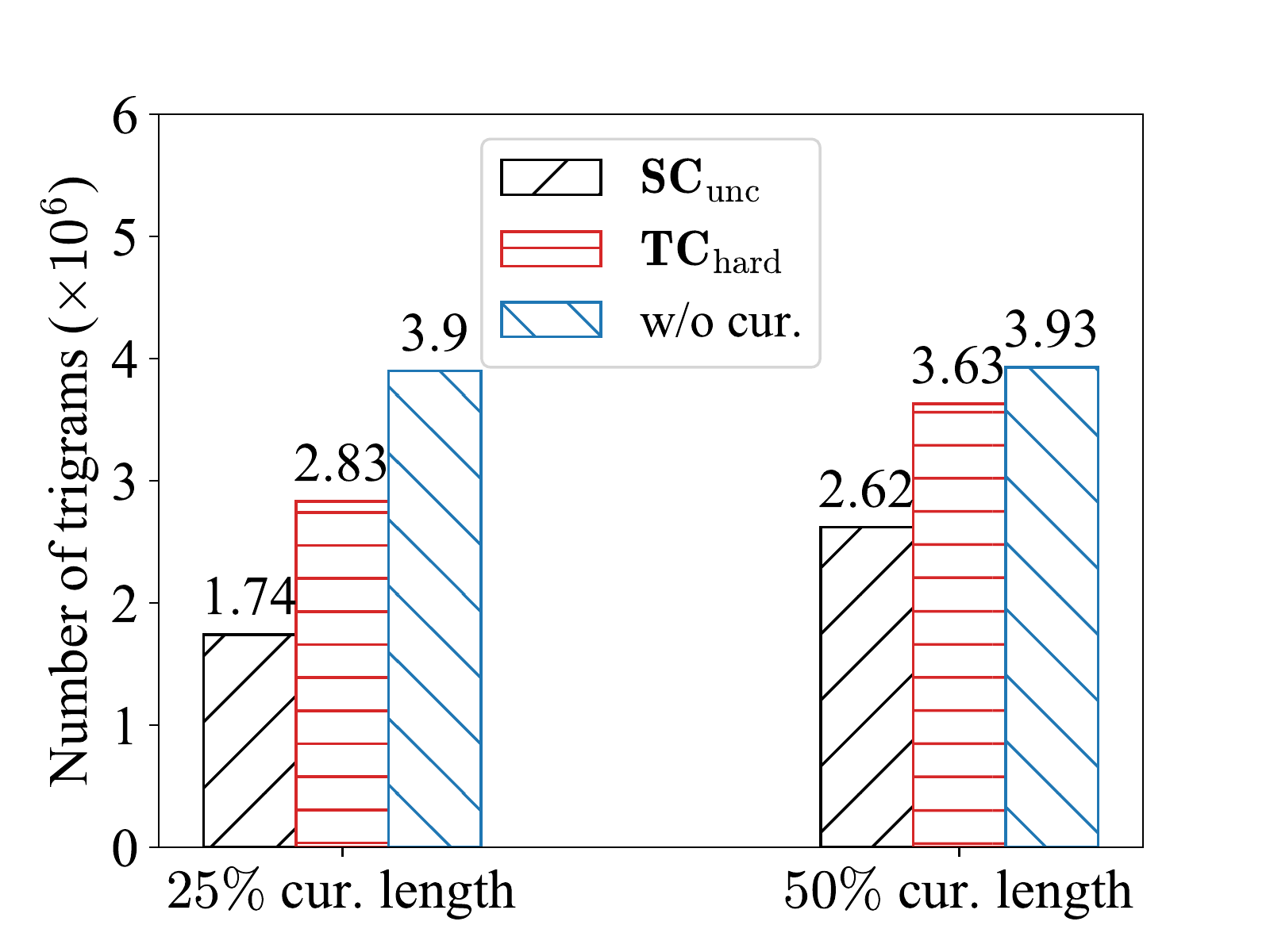}
    }
    \vspace{-0.2in}
	\caption{Sample diversity in early stage of training.}
	\label{fig:diversity_2}
	\vspace{-0.01in}
\end{figure}

\noindent $\bullet$ \textbf{Sample Diversity}. 
We compare the diversity of samples selected/created by $\textbf{SC}_{\text{unc}}$ and $\textbf{TC}_{\text{hard}}$ \footnote{Here, we consider $\textbf{TC}_{\text{hard}}$ as the diversity is easier to quantify.} on low-resource dataset De-En.
Recall that the samples selected by sentence-level curriculum is a subset of all sentence pairs. In contrast, the samples created by token-wise curriculum consist of all source sentences as well as the selected sub-sequences from all target sentences. Up till a fixed training iteration (e.g., $25\%$ of curriculum length), we measure the diversity by the number of unique trigrams summing over all selected/created sentences/sub-sequences.  As shown in Figure~\ref{fig:diversity_2}, the samples created by $\textbf{TC}_{\text{hard}}$ are more diverse at the early stage of training. 

\begin{figure*}[!ht]
    \centering
    \small
    \begin{tabular}{c@{}cc}
      \rotatebox[origin=c]{90}{Fr-En} & \raisebox{-0.5\height}{\includegraphics[width=0.48\textwidth]{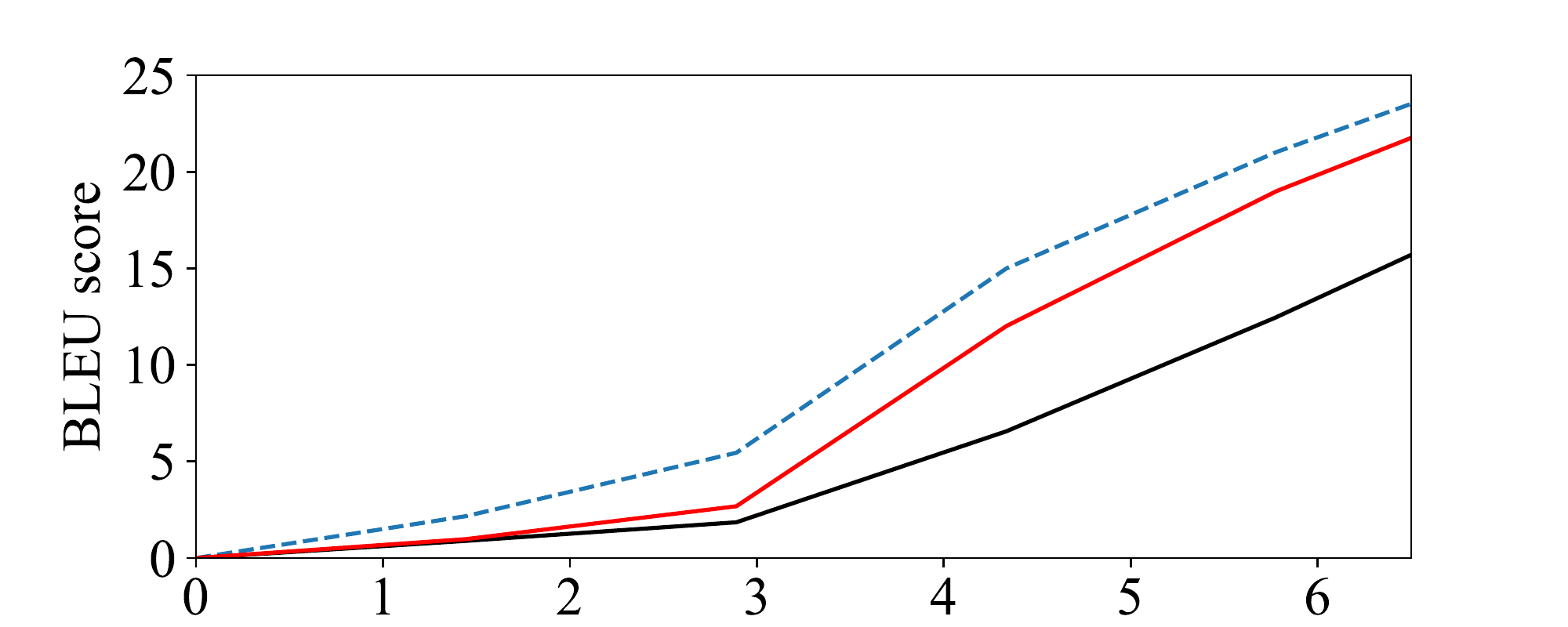}}  & \raisebox{-0.5\height}{\includegraphics[width=0.48\textwidth]{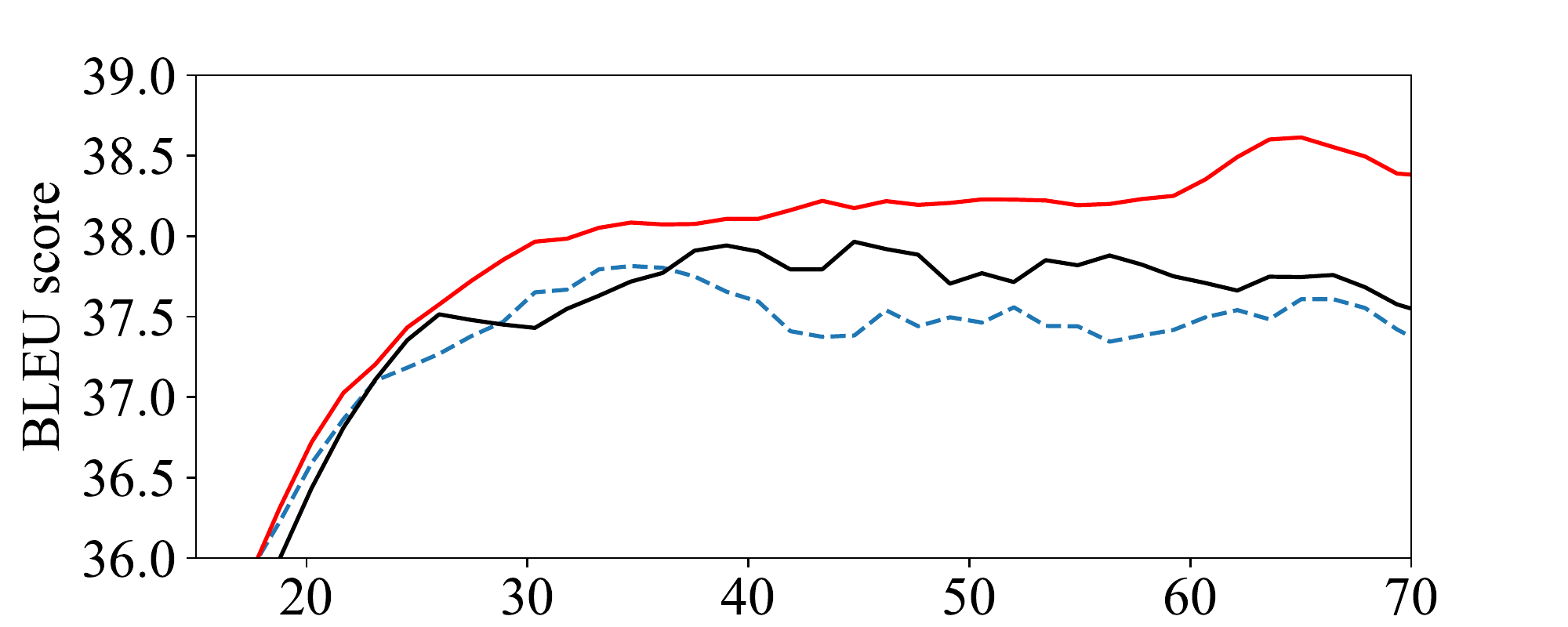}} \\
      \rotatebox[origin=c]{90}{En-De} & \raisebox{-0.5\height}{\includegraphics[width=0.48\textwidth]{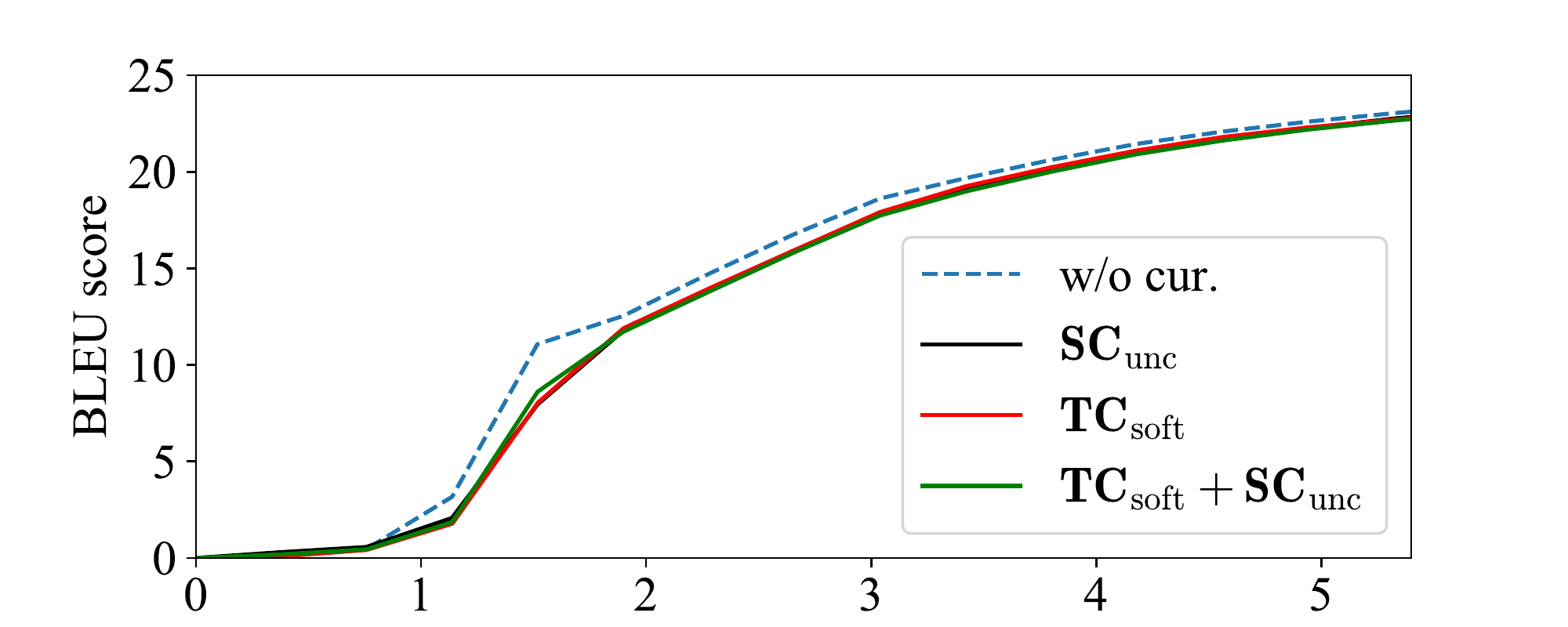}}  & \raisebox{-0.5\height}{\includegraphics[width=0.48\textwidth]{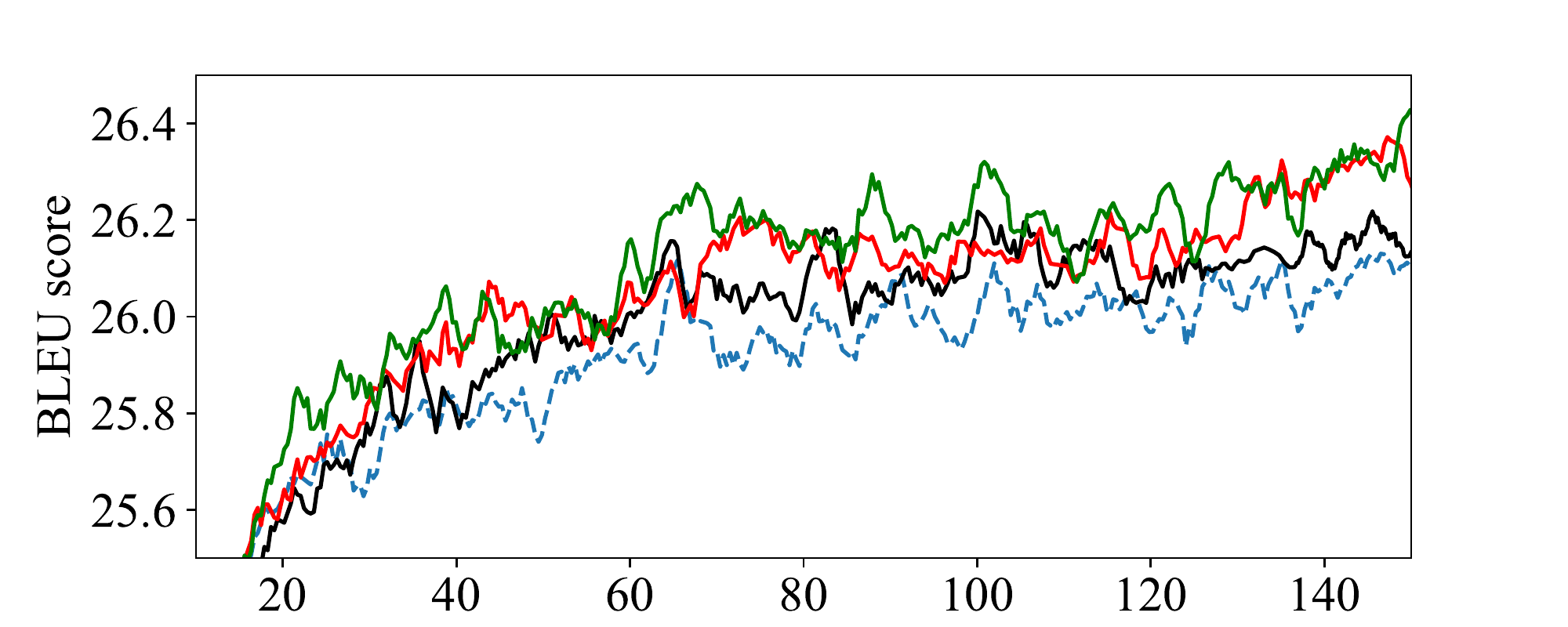}} \\
      & Number of updates ($\times 10^{3}$, During Curriculum)  &  Number of updates ($\times 10^{3}$, Till Convergence) \\
    \end{tabular}
    \caption{BLEU scores (dev) in low-resource dataset (Fr-En) and high-resource dataset (En-De). }
    \label{fig:low_high_2}
\end{figure*}

\noindent $\bullet$ \textbf{Learning Curve}. 
Figure~\ref{fig:low_high_2} shows the validation performance of $\textbf{SC}_{\text{unc}}$ and $\textbf{TC}_{\text{soft}}$ in both early and later stages of training. As can be seen, the BLEU score under the token-wise curriculum increases faster and more smoothly in the early stage. Furthermore, the model trained with the token-wise curriculum achieves a better generalization performance, while the model trained with sentence-level curriculum shows signs of over-fitting. We conjecture that such improvement comes from training with more diverse samples in the early stage.


\subsection{Ablation Study}
\label{sec:nmt_ablation}
We ablate some crucial designs of our curriculum, including the design of selecting consecutive tokens and the design of expanding the sub-sequence from beginning to the end of the sentence (referred as \textit{left-to-right}). We only consider $\textbf{TC}_{\text{hard}}$ in this section, as  $\textbf{TC}_{\text{soft}}$ is the improved version of $\textbf{TC}_{\text{hard}}$.

\noindent$\bullet$~\textbf{Consecutive Tokens vs. Random Tokens.} 
Here we study if selecting consecutive tokens is necessary for token-wise curriculum. It is natural to compare it with random sub-sequence curriculum, which uniformly samples the same number of tokens as $\textbf{TC}_{\text{hard}}$ (but not necessarily consecutive). Table~\ref{tab:consecutive} shows that the random curriculum does not show improvement upon the baseline. 

\begin{table}[!ht]
\begin{tabular}{l|ccc}
\toprule
\multicolumn{1}{l|}{}                 & \textbf{De-En} & \textbf{Fr-En} & \textbf{Ro-En}        \\ \midrule 
\multicolumn{1}{l|}{w/o Curriculum} & 34.33 & 37.21 & 32.10        \\ 
Random Sub-seq.                         & 34.46 & 37.30 & 32.20    \\
$\textbf{TC}_{\text{hard}}$             & 34.88 & 38.22 & 32.45        \\ \bottomrule 
\end{tabular}
\caption{BLEU score (test) comparison to random sub-sequence curriculum. }
\vspace{-5mm}
\label{tab:consecutive}
\end{table}

\noindent$\bullet$~\textbf{Teacher-forcing Loss vs. Beam Search Error Rate.} The \textit{left-to-right} design is motivated by the error accumulation of beam search decoding which makes the latter tokens more difficult to predict (Figure~\ref{fig:pre_ea}). 
Recall that, unlike beam search, the NMT models are trained in a teacher-forcing way. Therefore, we would like to know whether the teacher-forcing training loss can characterize sample difficulty. To answer this question, we select the sub-sequence with the lowest average teacher-forcing loss and the same number of tokens as $\textbf{TC}_{\text{hard}}$.
Table~\ref{tab:loss_aw} shows that the selection based on teacher-forcing loss outperforms the baseline, but does not work as well as the \textit{left-to-right} design. 

\begin{table}[!ht]
\begin{tabular}{@{\hskip2pt}l|ccc@{\hskip2pt}}
\toprule
\multicolumn{1}{l|}{}                         & \textbf{De-En}    & \textbf{Fr-En}    & \textbf{Ro-En} \\ \midrule 
\multicolumn{1}{l|}{w/o Curriculum }         & 34.33    & 37.21    & 32.10  \\
Subseq w/ Low. Loss                   &  34.63   & 37.53    & 32.29  \\
$\textbf{TC}_{\text{hard}} $            & 34.88    & 38.22    & 32.45  \\ \bottomrule
\end{tabular}
\caption{BLEU score (test) comparison to selecting based on sub-sequence average training loss.}
\label{tab:loss_aw}
\end{table}

\noindent$\bullet$~\textbf{Relative Positions of Sub-sequences}. 
We further explore whether choosing a sub-sequence expansion direction misaligned with the $\textit{left-to-right}$ decoding order can also improve the performance. We select initial sub-sequence not from the beginning of each sentence, instead, in the range of $30-40\%$, $60-70\%$ and $90-100\%$ of each sentence with the same expansion schedule. For example, by selecting the initial range as $90-100\%$, the expansion is in the $\textit{right-to-left}$ direction. By selecting the initial range as $30-60\%$, the sub-sequence is expanding bidirectionally. Table~\ref{tab:init_pos} shows that by choosing initial sub-sequence other than beginning of the sentence, the performance drops even below the baseline in some cases. This implies that the \textit{left-to-right} design is essential as it aligns with the decoding order.

\begin{table}[!ht]
\begin{tabular}{@{\hskip2pt}l|ccc}
\toprule
\multicolumn{1}{l|}{}                          & \textbf{De-En}    & \textbf{Fr-En}    & \textbf{Ro-En} \\ \midrule
\multicolumn{1}{l|}{w/o Curriculum }          & 34.33    & 37.21    & 32.10  \\ 
$\textbf{TC}_{\text{hard}}, 30-40\% $          & 34.08    & 37.48    & 32.25  \\
$\textbf{TC}_{\text{hard}}, 60-70\% $          & 34.17    & 37.12    & 32.03  \\
$\textbf{TC}_{\text{hard}}, 90-100\% $         & 34.18    & 37.45    & 32.43  \\ 
$\textbf{TC}_{\text{hard}}, 0-10\% $           & 34.88    & 38.22    & 32.45  \\ \hline
\end{tabular}
\caption{BLEU score (test) comparison to selecting initial sub-sequence from different relative positions.}
\label{tab:init_pos}
\end{table}

\vspace{-0.15in}
\section{Language Modeling Experiments}
\label{sec:lm_exp}
\vspace{-0.05in}
To demonstrate our token-wise curriculum can be applied to other sequence generation task, we presents experimental results on language modeling.  

\subsection{Data Preparation \& Processing}
\label{sec:lm_data} 
We conduct experiments on two popular word-level datasets: a preprocessed version of the Penn Treebank (PTB) \citep{mikolov2010recurrent} and the WikiText-2 (WT2) \citep{merity2016pointer}. PTB contains about $929$K training words, $73$K validation words, and $82$K test words. All capitalization, numbers and punctuation are removed as part of the preprocessing step. WT2 consists of around $2$M words extracted from Wikipedia articles. The dataset is lightly processed with capitalization, punctuation, and numbers retained. It is tokenized and preprocessed using the Moses \citep{koehn2007moses} with over $30$K vocabulary size.

\subsection{Model \& Training}
\label{sec:lm_model}
We use AWD-LSTM \citep{merityRegOpt}, a $3$-layer standard LSTM equipped with the drop-connection \citep{wan2013regularization} on recurrent weights. The model is trained with non-monotonically triggered averaged stochastic gradient descent (NT-ASGD), a variant of ASGD \citep{polyak1992acceleration}. We follow the training settings from \citet{merityRegOpt}\footnote{\small{\url{https://github.com/salesforce/awd-lstm-lm}}}and report performance in perplexity under static evaluation. 

We fix $\lambda_0$, $\gamma_0$ and $\alpha_0$ the same as in Section~\ref{sec:nmt_model}. The curriculum length $I$ is set to be $2100$ and $4200$ for PTB and WT2. See hyperparameters selection details in~\ref{app:param_study}.

\subsection{Main Results}
\label{sec:lm_main}
Table~\ref{tab:lm_main_results} shows the language modeling performance on PTB and WT2. As can be seen, both $\textbf{TC}_{\text{hard}}$ and $\textbf{TC}_{\text{soft}}$ outperform the baseline performance by over $0.5$ points of perplexity. Furthermore, $\textbf{TC}_{\text{soft}}$ slightly outperforms $\textbf{TC}_{\text{hard}}$ in both datasets.
\begin{table}[ht!]
\centering
\begin{tabular}{l|cc}
\toprule
                             & \textbf{PTB}         & \textbf{WT2}         \\\midrule  
w/o Curriculum                         & 58.96 & 65.54 \\ 
$\textbf{TC}_{\text{hard}}$      & 58.41 & 65.14 \\
$\textbf{TC}_{\text{soft}}$      & \textbf{58.23} & \textbf{65.09} \\ \hline
\end{tabular}
\caption{\label{tab:lm_main_results} The perplexity (test) in language modeling.}
\vspace{-3mm}
\end{table}
\section{Conclusion}

In this paper, we introduce a novel token-wise curriculum learning method for NMT. We show its superiority in low-resource setting, and is beneficial in high-resource setting. Different from existing works, we only consider a vanilla curriculum schedule, where the created sub-sequences expand linearly, as our focus is to validate the idea of token-wise design. We leave other potential scheduler design, e.g., training adaptive scheduler \citep{liu2020norm,xu2020curriculum}, as future discussion.


\newpage
\section*{Broader Impact}

This paper proposes a new curriculum learning method for training neural language models in sequence-to-sequence prediction tasks. Our designed curriculum neither introduces any social/ethical bias to the model nor amplify any bias in the data. We do not foresee any direct social consequences or ethical issues.

\bibliographystyle{acl_natbib}
\bibliography{naacl2021}
\clearpage

\appendix
\section{Appendix}
\label{sec:appendix}

\subsection{Datasets}
\label{app:datasets}
\noindent $\bullet$ \textbf{IWSLT14 De-En} We follow \citet{ott2019fairseq}\footref{fairseq} to split the dev/test sets.\\
\noindent $\bullet$ \textbf{IWSLT15 En-Vi} We use the standard TED dev2010 and tst2013 as dev and test set following \citet{platanios2019competence}. \\
\noindent $\bullet$ \textbf{IWSLT16 Fr-En} We use the standard TED tst2015 and tst2016 as dev and test set following \citet{platanios2019competence}. \\
\noindent $\bullet$ \textbf{WMT16 Ro-En} We use the standard newsdev-2016 and newstest-2016 are used as dev and test set.\\
\noindent $\bullet$ \textbf{WMT16 En-De} We use the standard newstest-2013 and newstest-2014 as dev and test set. 

\subsection{$\textbf{TC}$ Methods Implementation Details}
\label{app:training}
\noindent $\bullet$ \textbf{NMT Standard Training Experiments.} For all language pairs, we use a inverse square root schedule with weight decay rate of $1\times10^{-4}$, label smoothing ratio of $0.1$, and dropout rate of $0.3$. 

For low resource setting, we share the decoder and encoder output embeddings. We use dynamic batching with maximum tokens of $4096$ per GPU and train on $1$ GPU for $60$ epochs. 

For high resource setting, we share all the embeddings. We use dynamic batching with $14336$ tokens per GPU, accumulate gradient for $7$ steps, and train for $150$K updates. 

For extremely low-resource setting, we follow the same hyperparameter setting for each language pair.

\noindent $\bullet$ \textbf{NMT Transfer Learning Experiments.} We use 2 NVIDIA V100 GPUs for each experiment. We choose finetuning learning rate from $\{1\times 10^{-5}, 5 \times 10^{-5}, 5 \times 10^{-4}\}$. We use dynamic batch size, which is limited by GPU memory ($16$G per GPU). We report the evaluation results by conducting beam search with beam size of $5$ and length penalty of $0.6$ for datasets in WMT, and beam size of $5$ and length penalty of $2$ for datasets in IWSLT.

\subsection{$\textbf{TC}$ Methods Hyperparameter Selection}
\label{app:param_study}
\noindent $\bullet$ $\textbf{Selection of } \boldsymbol{\lambda}_0, \boldsymbol{\alpha}_0, \boldsymbol{\gamma}_0$. We choose $\lambda_0$ in $\{0.1, 0.2, 0.3\}$, $\gamma_0$ in $\{0.5, 0.6, 0.7, 0.8, 0.9\}$, and $\alpha_0$ in $\{12, 25, 37, 50\}$. We find that setting $\lambda_0$ in $\{0.1, 0.3\}$, $\gamma_0$ in $\{0.7, 0.9\}$ and $\alpha_0$ in $\{12, 25\}$ leads to less than $0.05$ variance in validation performance, suggesting $\textbf{TC}$ methods are insensitivity to hyperparameters. 

\noindent $\bullet$ $\textbf{Selection of }$ $\boldsymbol{I}$. In NMT experiments, we determine $I$ in a similar manner as \citet{platanios2019competence}: we train the baseline model and compute the number of training steps it takes to reach approximately $70\%$ of its final BLEU score. We then set $I$ to this value. In language modeling experiments, $I$ is determined similarly: we train the baseline model and set $I$ to be the number of training steps it takes to reach approximately $30\%$ initial perplexity + $70\%$ final perplexity.

\subsection{$\textbf{SC}$ Methods Implementation Details}
\label{app:baseline}
\noindent $\bullet$ $\textbf{SC}_{\text{r-sqrt}}$. We adopt the SR curriculum and $c_{sqrt}$ competence function setting in \citet{platanios2019competence}. We set initial competence $c_0$ to $0.01$ for all language pairs and set curriculum length $T$ in the same manner following \citet{platanios2019competence}. In addition, we adopt the special learning rate schedule as proposed in Equation (9) in the original paper, where we set $T_{\text{warmup}}=8000$.  

\noindent $\bullet$ $\textbf{SC}_{\text{norm}}$. Following \citet{liu2020norm}, we extract a word2vec embedding $E^{w2v}$ from a pre-trained Transformer-base model and measure sample difficulty on the source sentences embedding mapped through $E^{w2v}$. The initial competence $c_0$ is set to $0.01$ for all language pairs. For En-De, $\lambda_m$ and $\lambda_w$ are set to $2.5$ and $0.5$ following \citet{liu2020norm}. For low-resource datasets, we tune and choose $\lambda_m$ and $\lambda_w$ as $0.25$ and $0.05$, respectively.

\noindent $\bullet$ $\textbf{SC}_{\text{unc}}$. We follow \citet{zhou2020uncertainty} to use $4$ baby steps. We measure sample difficulty using the ``joint'' source and target uncertainty. It is obtained by evaluating the perplexity measured by a pre-trained 4-gram KENLM model \citep{heafield2011kenlm}).

\subsection{Validation Performance}
\label{app:valid}
\noindent $\bullet$ \textbf{NMT Standard Training Experiments.} Table~\ref{tab:valid_low} shows the validation performance on low-resource datasets. We report the BLEU score on the best checkpoint. and Table~\ref{tab:valid_high} shows the validation performance on a high-resource dataset En-De. We report the BLEU score on the averaged last $10$ checkpoints. We use the same beam search setting as in Section~\ref{sec:nmt_model}. 

\begin{table}[!htb]
\begin{tabular}{l|cccc}
\toprule
                                      & \textbf{En-Vi} & \textbf{De-En} & \textbf{Fr-En} & \textbf{Ro-En} \\ \midrule
w/o Cur.                              & 29.77          & 35.62 & 37.99   & 32.91        \\ 
$\textbf{SC}_{\text{r-sqrt}}$         & 29.75	       & 35.44 & 38.14	 & 32.95                                   \\
$\textbf{SC}_{\text{norm}}$           & 29.70          & 35.83 & 38.21   & 33.02                                  \\
$\textbf{SC}_{\text{unc}}$            & 29.32	       & 35.80 & 38.26	 & 32.75                                \\ \hhline{=====}
$\textbf{TC}_{\text{hard}}$           & 30.39          & 36.11 & 38.68   & 33.21       \\
$\textbf{TC}_{\text{soft}}$         & \textbf{30.42}   & \textbf{36.14} & \textbf{38.77}   & \textbf{33.24}            \\ \bottomrule
\end{tabular}
\caption{\label{tab:valid_low} The BLEU scores (dev) on low-resource datasets. }
\end{table} 

\begin{table}[!htb]
        \centering
        \begin{tabular}{l|c}
        \toprule
                                             & \textbf{En-De} \\ \midrule
        w/o Cur. ~ \citep{vaswani2017attention}  ~~~               & 26.10                    \\ 
        $\textbf{SC}_{\text{r-sqrt}}$  ~ \citep{platanios2019competence}     &  26.15                   \\
        $\textbf{SC}_{\text{norm}}$ ~    \citep{liu2020norm}     &   26.32                    \\
        $\textbf{SC}_{\text{unc}}$  ~ \citep{zhou2020uncertainty}     & 26.26                     \\ \hhline{==}
        $\textbf{TC}_{\text{hard}}$          &  26.44                     \\
        $\textbf{TC}_{\text{soft}}$          &  26.46\\ 
        $\textbf{TC}_{\text{soft}} + \textbf{SC}_{\text{norm}}$        & \textbf{26.57} \\ 
        $\textbf{TC}_{\text{soft}} + \textbf{SC}_{\text{unc}}$      & 26.48 \\ \bottomrule
        \end{tabular}
        \caption{\label{tab:valid_high} BLEU scores (dev) on a high-resource dataset. }
\end{table}

\noindent $\bullet$ \textbf{Language Modeling Experiments.}
Table~\ref{tab:lm_valid_results} shows the validation performance of the language modeling experiments. 
\begin{table}[t!]
\centering
\begin{tabular}{l|cc}
\hline
                             & \textbf{PTB}         & \textbf{WT2}         \\ \midrule
AWD-LSTM                         & 61.09 & 68.40 \\
$\textbf{TC}_{\text{hard}}$      & 60.85 & 68.32 \\
$\textbf{TC}_{\text{soft}}$      & 60.63 & 67.82 \\ \hline
\end{tabular}
\caption{\label{tab:lm_valid_results} The perplexity (dev) in language modeling.}
\end{table}

\subsection{Additional Analysis}
\label{app:add_ana}
To interpret how $\textbf{TC}$ benefits sequence generation, we further analyze whether $\textbf{TC}$ is able to alleviate the error accumulation. We conduct beam search with beam size of $5$ on Transformer-base model trained on De-En, and compute the averaged prediction error rate over the end $20\%$ tokens of the sentences. As shown in the Table~\ref{tab:err_acc_ana}, the model trained with $\textbf{TC}_{\text{hard}}$ suffers less from error accumulation than the model trained with $\textbf{TC}_{\text{hard}}$. In addition, $\textbf{TC}_{\text{hard}}$ particularly alleviates error accumulation in long sentences (i.e., sentences with length larger than $100$).

\begin{table}[!ht]
\begin{tabular}{@{\hskip2pt}l|ccc@{\hskip2pt}}
\toprule
\multicolumn{1}{l|}{}                   & \textbf{All Lengths} & \textbf{Lengths > 100} \\ \midrule
w/o Curriculum    & 76.8\%    & 99.3\%  \\
$\textbf{TC}_{\text{hard}}$              & 77.0\%    & 97.1\%   \\
$\textbf{TC}_{\text{hard}}$             & 75.7\%    & 87.7\%   \\ \bottomrule
\end{tabular}
\caption{Averaged prediction error rate over the end $20\%$ tokens for sentences in De-En dataset.}
\label{tab:err_acc_ana}
\end{table}
\end{document}